\pdfoutput=1
\documentclass[11pt,a4paper]{article}
\usepackage[hyperref]{eacl2021}
\usepackage{times}
\usepackage{latexsym}

\usepackage{graphicx}
\usepackage{tabularx}
\usepackage{soul}
\usepackage{times}
\usepackage{latexsym}
\usepackage{makecell}
\usepackage{arabtex}
\newcommand\ararab[2][]{{\setcode{utf8}\RL{#2}}}
\usepackage{utf8}
\usepackage{graphicx}
\usepackage{url}
\usepackage{paralist}
\usepackage{qtree}
\usepackage{color, colortbl}

\usepackage{epstopdf}
\usepackage[latin1]{inputenc}
\usepackage{tipa}
\usepackage[T1]{fontenc}
\usepackage{hyperref}
\usepackage{xstring}
\newcommand{\hide}[1]{}

\newcommand{\TAMARBUTA}{{$\hbar$}}

\newcommand{\AYN}{{$\varsigma$}}

\usepackage{pifont}
\newcommand{\cmark}{\ding{51}}

\usepackage{microtype}

\aclfinalcopy 

\title{Automatic Romanization of Arabic Bibliographic Records}

\author{Fadhl Eryani \textmd{and} Nizar Habash \\
Computational Approaches to Modelling Language (CAMeL) Lab\\
 New York University Abu Dhabi, UAE \\
 \texttt{\{fadhl.eryani,nizar.habash\}@nyu.edu}}
 
\date{}

\begin{document}

\maketitle

\begin{abstract}
    International library standards require cataloguers to tediously input Romanization of their catalogue records for the benefit of library users without specific language expertise.  In this paper, we present the first reported results on the task of automatic Romanization of {\it undiacritized} Arabic bibliographic entries. This complex task requires the modeling of Arabic phonology, morphology,  and even semantics. We collected a 2.5M word corpus of parallel Arabic and Romanized bibliographic entries, and
    benchmarked  a number of models that vary in terms of complexity and resource  dependence. Our best system reaches 89.3\% exact word Romanization on a blind test set.  We make our data and code publicly available. 

\end{list}
\end{abstract}




\section{Introduction}
\setcode{utf8}
\vocalize

\begin{figure}[]
\centering
 \includegraphics[width=0.48\textwidth]{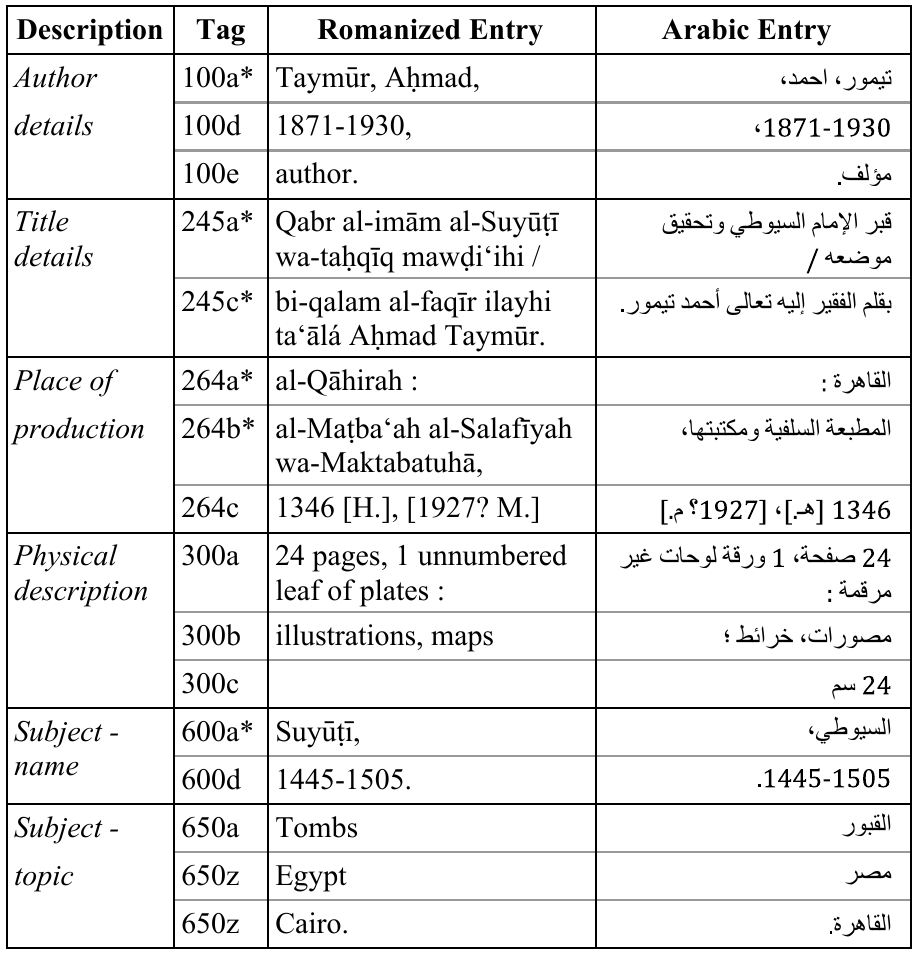}
  \caption{A bibliographic record for \newcite{Taymur:1927:qabr} in Romanized and original Arabic forms. }
\label{loc-example}
\end{figure}

Library catalogues comprise a large number of bibliographic records consisting of entries that provide specific descriptions of library holdings. Records for Arabic and other non-Roman-script language materials ideally include 
Romanized entries to help researchers without language expertise, e.g., Figure~\ref{loc-example}.
There are many Romanization standards such as the ISO standards used by French and other European libraries, and the ALA-LC (American Library Association and Library of Congress) system \cite{LOC:2017:romanization} widely adopted by North American and UK affiliated libraries. 
These Romanizations are applied manually  by librarians across the world -- a tedious error-prone task.

In this paper, we present, to our knowledge, the first reported results on automatic Romanization of {\it undiacritized} Arabic bibliographic entries. This is a non-trivial task as it requires modeling of Arabic phonology, morphology and even semantics.
We collect and clean a 2.5M word corpus of parallel Arabic and Romanized bibliographic entries, and evaluate a number of 
models that vary in terms of complexity and resource dependence. Our best system reaches 89.3\% exact word Romanization on a blind test set. We make our data and code publicly available for researchers in Arabic NLP.\footnote{\url{https://www.github.com/CAMeL-Lab/Arabic\_ALA-LC\_Romanization}}
 
\section{Related Work} 

\paragraph{Arabic Language Challenges}
Arabic poses a number of challenges for NLP
in general, and the task of Romanization in particular.
Arabic is morphologically rich, uses a number of clitics, and is written using an Abjad script with optional diacritics, all leading to a high degree of ambiguity.
The Arabic script does not include features such as capitalization which is helpful for NLP in a range of Roman script languages.
There are a number of enabling technologies for Arabic that can help, e.g., 
MADAMIRA \cite{Pasha:2014:madamira},
 Farasa \cite{Abdelali:2016:farasa},
and CAMeL Tools \cite{Obeid:2020:cameltools}.
In this paper we use MADAMIRA to provide diacritics, morpheme boundaries and English gloss capitalization information as part of a rule-based Romanization technique. 

\paragraph{Machine Transliteration}
{\it Transliteration} refers to the mapping of text from 
one script to another. 
Romanization is specifically transliteration into the Roman script \cite{Beesley:1997:romanization}.
There are many ways to transliterate and Romanize, varying in terms of detail, consistency, and usefulness. 
Commonly used name transliterations \cite{Al-Onaizan:2002:machine} and so-called Arabizi transliteration \cite{Darwish:2014:arabizi} tend to be lossy and inconsistent
while strict orthographic transliterations such as Buckwalter's \cite{Buckwalter:2004:buckwalter}
tend to be exact but not easily readable. 
The ALA-LC transliteration is a relatively easy to read standard that requires a lot of details on phonology, morphology and semantics.
There has been a sizable amount of work on 
mapping Arabizi to Arabic script using a range of techniques from rules to neural models
\newcite{chalabi-gerges-2012-romanized, Darwish:2014:arabizi,Al-Badrashiny:2014:automatic,guellil2017arabizi,YOUNES2018238, Shazal:2020:unified}.
%
In this paper we make use of a number of insights and techniques from work on Arabizi-to-Arabic script transliteration, but apply them in the opposite direction to map from Arabic script to
a complex, detailed and strict Romanization. We compare rule-based and corpus-based techniques including a Seq2Seq model based on the publicly available code base of 
\newcite{Shazal:2020:unified}.

\section{Data Collection} 
\label{datasets}
%



\begin{table}[t!]
\centering
\begin{tabular}{|l|r|r|c|}
\hline
\textbf{Split} & \multicolumn{1}{c|}{\textbf{Bib Records}} & \textbf{Entries} & \textbf{Words} \\ \hline\hline
\textbf{Train} & 85,952 (80\%) & 479,726     & $\sim$2M   \\\hline
\textbf{Dev}  & 10,744 (10\%) & 59,964      & $\sim$250K    \\\hline
\textbf{Test} & 10,743 (10\%) & 59,752      & $\sim$250K    \\\hline\hline
\bf Total & 107,439 \verb|  |   & 599,442     & $\sim$2.5M   \\\hline
\end{tabular}
\caption{Corpus statistics and data splits.}
\label{datsets}
\end{table}

\paragraph{Sources} We collected bibliographic records from three publicly available xml dumps stored in the machine-readable cataloguing (MARC) standard, an international standard for storing and describing bibliographic information. The three data sources are the Library of Congress
(LC) (10.5M), the University of Michigan
(UMICH) (680K), and New York University Abu Dhabi's Arabic Collections Online
(ACO) (12K), amounting to 11.2 million records in total.

\paragraph{Extraction} 
From these collections, we extracted 107,493 records
that 
are specifically tagged with the Arabic language code (MARC 008 ``ara''). 

\paragraph{Filtering} Within the extracted records we filter out some of the entries using two strategies. First, we
used a list of 33 safe tags (determined using their definitions and with empirical sampling check)
to eliminate all entries that include a mix of translations, control information, and dates. The star-marked tags in Figure~\ref{loc-example} are all included, while the rest are filtered out. Second, we eliminated all entries with mismatched numbers of tokens. This check was done after a cleaning step that corrected for common errors and inconsistencies in many entries such as punctuation misalignment and incorrect separation of the conjunction \<و+> {\it wa+}\footnote{Strict orthographic transliteration using the HSB scheme \cite{Habash:2007:arabic-transliteration}. } `and' clitic. As a result of this filtering, a small number of additional records are eliminated since all their entries were eliminated. 
The total number of retained records is 107,439.
The full details on extraction and filtering are provided as part of the project's public github repo (see footnote 1).

\paragraph{Data Splits}
Finally, we split the remaining collection of records into Train, Dev, and Test sets. Details on the number of records, entries, and words they contain is presented in Table~\ref{datsets}. We make our data and data splits available (see footnote 1).
 


\section{Task Definition and Challenges}
As discussed above, there are numerous ways to ``transliterate'' from one script to another. 
In this section we focus on the
Romanization of undiacritized Arabic bibliographic entries into the ALA-LC standard. 
Our intention is to highlight the important challenges of this task in order to justify the design choices we make in our approaches. 
For a detailed reference of the ALA-LC Arabic Romanization standard, see \cite{LOC:2012:Arabic}.
 
\paragraph{Phonological Challenges} While Romanizing Arabic consonants is simple, the main challenge is in identifying unwritten phonological phenomena, e.g., short vowels, under-specified long vowels, consonantal gemination, and nunnation, all of which require modeling Arabic diacritization.  

\paragraph{Morphosyntactic Challenges}
Beyond basic diacritization modeling, the task requires some morphosyntactic modeling: examples include
(a) proclitics such as the definite article, prepositions and conjunctions are marked with a hyphen,
(b) case endings are dropped, except before pronominal enclitics, 
(c) the silent Alif, appearing in some masculine plural verbal endings, is ignored,
and (d) the Ta-Marbuta ending can be written as {\it h} or {\it t} depending on the morphosyntactic state of the noun. For more information on Arabic morphology, see \cite{Habash:2010:introduction}.
 
\paragraph{Semantic Challenges}
Proper nouns need to be marked with capitalization on their first non-clitic alphabetic letter. Since Arabic script does not have ``capitalizations'', this effectively requires named-entity recognition.
The Romanization of the word \<القاهرة>
{\it AlqAhr{\TAMARBUTA}} `Cairo' as {\it al-Q\=ahirah}
in Figure~\ref{loc-example} illustrates elements from all challenge types.

\paragraph{Special Cases}
The Arabic ALA-LC guidelines include a number of special cases, e.g.,
the word \<بن>
{\it bn} `son~of' is Romanized as {\it ibn}, and proper noun 
 \<عمرو>
{\it {\AYN}mrw} is Romanized as {\it `Amr}.


\begin{table*}[ht!]
 \centering
\setlength{\tabcolsep}{5pt} 
  \begin{tabular}{|l||r|c|c||r|r|r||r|r|r|}
  \hline
\textbf{} & \multicolumn{1}{c|}{\textbf{Corpus }} & \textbf{Morph}&\textbf{Char} &\multicolumn{3}{c||}{\textbf{Dev}} &\multicolumn{3}{c|}{\textbf{Test}}  \\ \cline{5-10}
\textbf{Model} & \multicolumn{1}{c|}{\textbf{Size }}& \textbf{Trans}&\textbf{Trans} 

& \multicolumn{1}{c|}{\textbf{Exact}}& \multicolumn{1}{c|}{\textbf{CI}}& \multicolumn{1}{c||}{\textbf{CPI}}
& \multicolumn{1}{c|}{\textbf{Exact}}& \multicolumn{1}{c|}{\textbf{CI}}& \multicolumn{1}{c|}{\textbf{CPI}} \\
  \hline\hline
\bf Rules Simple & 0 & & \cmark
 & 16.2 &     17.4 &     17.8    &     16.1 &     17.3 &     17.7 \\ \hline
\bf Rules Morph           & 0 & \cmark & \cmark &     67.4 &     83.5 &     84.8  &     67.4 &     83.6 &     84.9 \\ \hline\hline
\bf MLE Simple 1/64       & 31K & & \cmark &63.6&69.8&     71.1 & \multicolumn{3}{c}{} \\ \cline{1-7}
\bf MLE Simple 1/32       & 63K & & \cmark&     68.5 &     75.1 &     76.4  & \multicolumn{3}{c}{} \\ \cline{1-7}
\bf MLE Simple 1/16       & 125K & & \cmark&     73.0 &     79.9 &     81.3   & \multicolumn{3}{c}{}\\ \cline{1-7}
\bf MLE Simple 1/8        & 250K & & \cmark       &     75.6 &     82.8 &     84.2  & \multicolumn{3}{c}{}\\ \cline{1-7}
\bf MLE Simple 1/4        & 500K & & \cmark&     80.3 &     87.2 &     88.6  & \multicolumn{3}{c}{}\\ \cline{1-7}
\bf MLE Simple 1/2        & 1M & & \cmark       &     82.7 &     89.5 &     90.9  & \multicolumn{3}{c}{}\\ \hline
\bf MLE Simple            & 2M & & \cmark&84.0&     90.7 &     92.1   &     84.1 &     90.8 &     92.2 \\ \hline
\bf MLE Morph             & 2M & \cmark      & \cmark&84.7&     91.6 &     93.0   &     84.8 &     91.7 & \bf 93.2 \\ \hline\hline

\bf Seq2Seq 1/64          & 31K & &&     6.3  &     7.5  &     10.2& \multicolumn{3}{c}{} \\ \cline{1-7}
\bf Seq2Seq 1/32          & 63K & &&     28.3 &     31.0 &     38.1 & \multicolumn{3}{c}{}\\ \cline{1-7}
\bf Seq2Seq 1/16          & 125K & &&     64.9 &     69.1 &     70.5& \multicolumn{3}{c}{} \\ \cline{1-7}
\bf Seq2Seq 1/8           & 250K & &&     75.5 &     79.6 &     80.9& \multicolumn{3}{c}{} \\ \cline{1-7}
\bf Seq2Seq 1/4           & 500K & &&     82.5 &     85.8 &     87.1& \multicolumn{3}{c}{} \\ \cline{1-7}
\bf Seq2Seq 1/2           & 1M & &&     85.9 &     88.6 &     90.1 & \multicolumn{3}{c}{}\\ \hline
\bf Seq2Seq               & 2M & &&87.2&     89.7 &     90.9      &     87.3 &     89.8 &     91.0 \\ \hline

\bf Seq2Seq + Rules Morph & 2M & \cmark & \cmark&88.8&     91.6 &     92.9   &     88.9 &     91.7 &     93.0 \\ \hline

\bf Seq2Seq + MLE Simple  & 2M & & \cmark& \bf 89.2 & \bf 91.8 & \bf 93.1    & \bf 89.3 & \bf 91.9 & \bf 93.2 \\ \hline

\bf Seq2Seq + MLE Morph   & 2M & \cmark      & \cmark&\bf 89.2 & \bf 91.8 & \bf 93.1   & \bf 89.3 & \bf 91.9 & \bf 93.2 \\ \hline

\end{tabular}
  \caption{Dev and  Test Romanization word accuracy (\%). (CI = case-insensitive, and CPI = case and punctuation-insensitive)} 
  \label{results} 
  \end{table*}

\section{Romanization Models}
\label{techniques}

We compare multiple Romanization models built using four basic techniques with different expectation about training data availability, contextual modeling, and system complexity.  The models are listed in Table~\ref{results}.

\paragraph{CharTrans Technique} 
Our baseline technique is an extremely simple character transliteration approach utilizing regular expressions and exception lists.
This technique is built based on the ALA-LC guidelines, and is inspired by the work of  \newcite{Biadsy:2009:improving}; it comprises 104 regex, 13 exceptions, and one capitalization rule (for entry-initial words). This technique accepts diacritized, undiacritized or partially diacritized input. Model {\bf Rules Simple} uses CharTrans only.

\paragraph{MorphTrans Technique} 
This technique relies on the morphological disambiguation system MADAMIRA \cite{Pasha:2014:madamira} to provide diacritization, morpheme boundaries, POS tags and English glosses for the Arabic  input.
Morpheme boundaries are used to identify clitic hyphenation points. POS tags and capitalization in English glosses are used to decide on what to capitalize in the transliteration.  We strip diacritical morphological case endings, but keep other diacritics.  We utilize the CharTrans technique to finalize the Romanization starting with the diacritized, hyphenated and capitalization marked words.  For words unknown to the morphological analyzer, we simply back off to the CharTrans technique. Model {\bf Rules Morph} uses MorphTrans with CharTrans backoff.  

\paragraph{MLE Technique} 
Unlike the previous two techniques, MLE (maximum likelihood estimate) relies on the parallel training data we presented in Section~\ref{datasets}.  This simple technique works on white-space and punctuation tokenized entries and learns simple one-to-one mapping from Arabic script to Romanization. The most common Romanization for a particular Arabic script input in the training data is used. The outputs are detokenized to allow strict matching alignment with the input. Faced with OOV (out of vocabulary), we back off to the MorphTrans technique (Model {\bf MLE Morph}) or CharTrans Technique (Model {\bf MLE Simple}). In Table~\ref{results}, we also study the performance of {\bf MLE Simple} with
different corpus sizes.

\paragraph{Seq2Seq Technique} 
Our last technique also relies on existing training data. We use
an encoder-decoder character-level sequence-to-sequence architecture 
closely following  \newcite{Shazal:2020:unified} (although in reverse direction). 

The encoder consists of two gated recurrent unit (GRU) layers \cite{Cho:2014:learning} with only the first layer being bidirectional,
and the decoder has two GRUs with attention \cite{Luong:2015:effective}. 
For the input, we used character embeddings concatenated with embeddings of the words in which the characters appear. 
For all other setting details, see \newcite{Shazal:2020:unified}'s Line2Line model.  We also show how \textbf{Seq2Seq} performs with different corpus sizes in Table \ref{results}.
%
%

The Seq2Seq technique is known for occasionally dropping tokens, which in our case leads to misalignment with the Arabic input. 
To handle this issue in model {\bf Seq2Seq}, we align its output and fill such gaps using  the outputs produced by three other techniques, thus creating models {\bf Seq2Seq+Rules Morph}, {\bf Seq2Seq+MLE Simple}, and {\bf Seq2Seq+MLE Morph}. The alignment technique we use relies on minimizing character-edit distance between present words to identify missing ones.

\paragraph{Comparing the Techniques} 
The CharTrans and MorphTrans techniques do not need parallel data, while the MLE and Seq2Seq techniques do.  Furthermore, the MorphTrans and Seq2Seq techniques make use of available context: in MorphTrans, we use context-aware monolingual morphological disambiguation; and in Seq2Seq we model parallel examples in context. In contrast, neither the MLE technique nor the CharTrans technique use the context of the words being mapped.



%

\begin{table*}[th!]
\centering
\begin{tabular}{|c|l|r|r|l|l|}
\hline
\multicolumn{2}{|c|}{\bf Error Type} & \bf Counts & \multicolumn{1}{l|}{\bf Source} & \bf Prediction & \bf Target \\\hline \hline
\bf Gold &  \bf  Romanization   & 34 & \<إبراهيم> & Ibr\=ah\={\i}m    & Ibr\=ahim          \\
52 &      &        & \<الأشقر>      & al-Ashqar       & Ashqar           \\
    & &       & \<ندوات>       & Nadaw\=at         & Nadw\=at           \\\cline{2-6}
& \bf Alignment  & 8      & \<أحمد بو حسن.> & A\textsubdot{h}mad  B\=u \textsubdot{H}asan. & B\=u  \textsubdot{H}asan,  A\textsubdot{h}mad.\\ \cline{2-6}
   & \bf  Source & 5      & \<الطبع>       & al-\textsubdot{T}ab` & al-\textsubdot{T}ab`ah     \\  \cline{2-6}
   & \bf Translation  & 5      & \<شعر.>        & shi`r. & Poems. \\ \hline \hline

\bf System    & \bf Romanization& 36     & \<الريف>       & al-Rayf & al-r\={\i}f           \\
48 & && \<حدث>&\textsubdot{h}adath	 & \textsubdot{H}addatha\\
& &&\<خسارة،> & Khass\=arah, &	Khas\=arah.\\\cline{2-6}
& {\bf Hallucination}& 10     & \<الادارية.>   & Ta\textsubdot{s}al-Id\=ar\={\i}yah. & al-Id\=ar\={\i}yah.  \\ \cline{2-6}
&  \bf Valid variant & 2      & \<السوفياتي>  & al-S\=ufy\=at\=\i      & al-S\=ufiy\=at\=\i     \\ 

\hline
\end{tabular}
\caption{Error types, counts, and examples on a sample of 100 \textbf{Seq2Seq+MLE Morph} predictions.}
\label{error_analysis} 
\end{table*}

\section{Experimental Results}
Table~\ref{results} presents the Dev and Test results for the models discussed in the previous section.  
All results are in terms of three word accuracy metrics: exact match (Exact), case-insensitive match (CI), and case and punctuation-insensitive match (CPI).  

The {\bf Rules Simple} baseline manages to correctly produce an exact answer in close to 1/6th of all the cases.
{\bf Rules Morph}, which uses no training data, misses about 1/3rd of all exact transliteration matches; however, about half of the errors are from capitalization issues. 

The {\bf MLE Simple} with 2M words cuts the error from Rules Morph by 51\% (Exact) and 44\% (CI).  
Notably {\bf Rules Morph} outperforms {\bf MLE Simple} with 31K words in Exact match, and  {\bf  MLE Simple} with 250K words in CI match.

The {\bf  MLE Morph} model improves over {\bf MLE Simple} by  $\sim$1\% absolute in all metrics (5\% and 10\% error reduction in Exact and CI, respectively). 

The {\bf Seq2Seq} model outperforms the {\bf MLE Morph} model by 2.5\% absolute (16\% error reduction) in Exact match, but under-performs in CI match.  
The \textbf{Seq2Seq} performance is comparatively much poorer with less data. With 31K words,
\textbf{MLE Simple}'s performance is 10 times better than \textbf{Seq2Seq}; and their performance only becomes comparable with 250K words.

We observe that $\sim$2\% of the {\bf Seq2Seq}  output words are missing, contributing negatively to the system's results. Of the three models that address this issue through alignment and combination, {\bf Seq2Seq+Rules Morph}, {\bf Seq2Seq+MLE Simple}, and {\bf Seq2Seq+MLE Morph}, the last two using the MLE technique are the best performers overall in Exact match. It's noteworthy that in CPI match, \textbf{MLE Morph}'s performance is almost equivalent to the best systems' performance.

The CPI metric values are consistently higher than CI by $\sim$1.3\% absolute for all models.

Blind test results presented in the right hand side of Table~\ref{results} are consistent with Dev results.
 

\section{Error Analysis}

We classified a sample of 100 word errors (ignoring capitalization and punctuation) from the Dev set of our best performing model ({\bf Seq2Seq+MLE Morph}). Our classification results are presented in Table~\ref{error_analysis} along with representative examples.

\paragraph{Gold Errors} We found 52 gold errors, where the human-provided target reference is incorrect. 
Romanization errors such as typos, incorrect vowelization, and dropped definite articles, constitute roughly 65\% of
gold errors. The rest of the errors include issues such as first and last name flipping which we classify as an alignment issue,  Arabic input source typos, and errors in which the target is a translation instead of a Romanization.
Notably, we observe that our {\bf Seq2Seq+MLE Morph} model generates correct predictions for 85\% of all gold error cases.


\paragraph{System Errors} Romanization errors make up 75\% of system errors. The vast majority of these mistakes are due to wrong prediction of vowels or gemination.  
An additional 21\% of the errors is due to Seq2Seq model hallucinations of characters unsupported by the source input.
We also encountered 2 predictions that did not match the target reference but are correct variants. In $\sim$44\% of system error cases,   outputs  generated by the {\bf MLE Morph} or {\bf Rules Morph} models are in fact correct, but were not chosen during alignment and combination because of existing Seq2Seq answers.  


\section{Conclusions and Future Work} 

We presented a new task for Arabic NLP, namely the Romanization of Arabic bibliographic records. Our extracted corpus and benchmark data splits, as well as our code base will be publicly available. 

In the future, 
we plan to create an online Romanization interface to assist librarians.
As more data is created  efficiently, better models can be created. 

We also plan to exploit the latent annotations in bibliographic records
for improving Arabic NLP tools, e.g. using vowelization for automatic diacritization and possible morphological disambiguation \cite{Habash:2016:exploiting}, marked clitics for 
tokenization, and  Roman-script capitalization for 
Arabic named entity recognition.

\section*{Acknowledgments} 
This work was carried out on the High Performance Computing resources at New York University Abu Dhabi (NYUAD).
We thank Salam Khalifa, Ossama Obeid, Justin Parrott, and Alexandra Provo for helpful conversations. 
And we especially thank Elie Kahale for introducing us to this interesting challenge during the Winter Institute in Digital Humanities at NYUAD.

\bibliography{camel-bib-v2,extra}
\bibliographystyle{acl_natbib}

\end{document}